\pdfoutput=1
\documentclass[11pt]{article}

\usepackage[review]{acl}

\usepackage{times}
\usepackage{latexsym}

\usepackage[T1]{fontenc}
\usepackage[utf8]{inputenc}

\usepackage{microtype}

\usepackage{amsfonts}
\usepackage{amsmath}
\usepackage{booktabs}
\usepackage{multirow}
\usepackage{graphicx}
\usepackage{enumitem}
\usepackage{textcomp}
\usepackage{booktabs}
\usepackage{subfigure}
\usepackage{array}

\title{HC3 Plus: A Semantic-Invariant Human ChatGPT Comparison Corpus}

\author{Zhenpeng Su\textsuperscript{\rm 1,2}\footnotemark[1], Xing Wu\textsuperscript{\rm 1,2}\footnotemark[1], Wei Zhou\textsuperscript{\rm 1,2}\footnotemark[2], Guangyuan Ma\textsuperscript{\rm 1,2}, Songlin Hu\textsuperscript{\rm 1,2}\footnotemark[2] \\
  \textsuperscript{\rm 1}Institute of Information Engineering, Chinese Academy of Sciences, Beijing, China\\
  \textsuperscript{\rm 2}School of Cyber Security, University of Chinese Academy of Sciences, Beijing, China\\
  \texttt{\{suzhenpeng,wuxing,zhouwei,maguangyuan,husonglin\}@iie.ac.cn} \\}
\begin{document}
\maketitle
\begin{abstract}
ChatGPT has garnered significant interest due to its impressive performance; however, there is growing concern about its potential risks, particularly in the detection of AI-generated content (AIGC), which is often challenging for untrained individuals to identify. Current datasets used for detecting ChatGPT-generated text primarily focus on question-answering tasks, often overlooking tasks with semantic-invariant properties, such as summarization, translation, and paraphrasing.
In this paper, we demonstrate that detecting model-generated text in semantic-invariant tasks is more challenging. To address this gap, we introduce a more extensive and comprehensive dataset that incorporates a wider range of tasks than previous work, including those with semantic-invariant properties. In addition, instruction fine-tuning has demonstrated superior performance across various tasks. In this paper, we explore the use of instruction fine-tuning models for detecting text generated by ChatGPT.

\end{abstract}

\section{Introduction}
In recent years, artificial intelligence has made significant advancements. Since the release of ChatGPT, large-scale language models (LLMs) fine-tuned on GPT-3.5 using Reinforcement Learning from Human Feedback (RLHF) have garnered extensive attention. ChatGPT has demonstrated strong performance across various tasks and excels at generating human-like text. As shown in ~\cite{DBLP:journals/corr/abs-2301-07597}, the Turing test reveals that it is challenging for individuals unfamiliar with ChatGPT to distinguish between texts generated by ChatGPT and those written by humans.
\begin{figure}[h]
    \centering
    \includegraphics[width=0.48\textwidth]{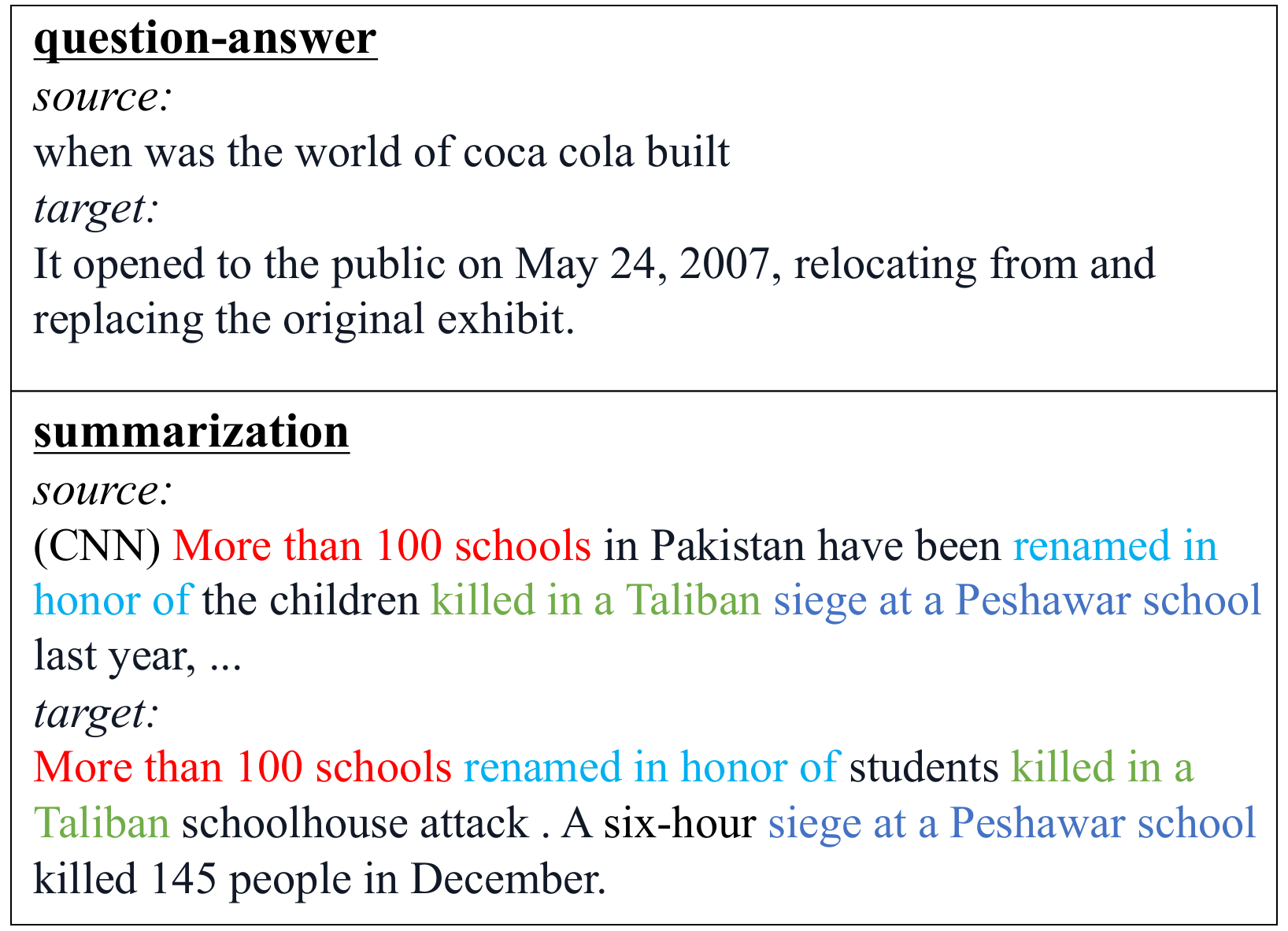}
    \caption{Examples of QA and summarization, the same color indicates fragments that appear in both source and target sentences.}
    \label{fig:compare}
\end{figure}

As more and more AI-generated content appears on social media, it could increase the risk of AI-generated disinformation and harmful text spreading due to the difficulty for humans to distinguish them~\cite{DBLP:journals/corr/abs-2301-07597}. Therefore, how to detect the text generated by the large language model has become a recent research hotspot. In fact, a good detector can be used for information supervision and accountability, making the source of information clearer. 
~\citet{DBLP:journals/corr/abs-2301-07597} introduced HC3, a dataset containing both ChatGPT-generated text and human-written text, designed for training AI text detection models. They first collected nearly $40,000$ questions and responses from human experts through social media and wiki text. ChatGPT was then used to generate answers to these questions. Using this collected data, they established two detection methods: training a logistic regression model based on GLTR~\cite{DBLP:conf/acl/GehrmannSR19} for classification and using RoBERTa~\cite{DBLP:journals/corr/abs-1907-11692} to train a binary classification model.

While their detector excels in performance on their test set, it is primarily focused on the question-answering (QA) task. In the QA task, the responses produced by the language model merely need to answer the requested question, without strict adherence to the semantics of the raw question. However, for tasks such as summarization, translation, and paraphrasing, the LLM must carefully consider the semantic nuances of the input sentences. 
The generated sentences must stay true to the meaning of the raw sentences without deviating from them. In other words, the output sentences need to maintain semantic invariance with the raw sentences. At the same time, the raw sentence used as input can be used as a reference for the output of the LLM. 
In the example shown in Figure \ref{fig:compare}, when comparing summarization with QA, there is an amount of overlapping vocabulary between the source sentence and the target sentence, and they have similar semantics. That may make it more difficult to recognize that the text is generated by the LLM. 

\begin{table}[!ht]
\resizebox{\linewidth}{!}{
\begin{tabular}{l|cc|cc|c}
\toprule 
Dataset & \multicolumn{2}{c|}{Human} & \multicolumn{2}{c|}{ChatGPT}  & ~ \\
~ & Precision & Recall & Precision & Recall & Accuracy  \\
\midrule
     \textbf{CNN/DailyMail} & 0.63 & 0.99 & 0.99 & 0.42 & 0.71 \\
     \textbf{LCSTS} & 0.52 & 0.99 & 0.88 & 0.09 & 0.54 \\
     \textbf{WMT'19 de$\rightarrow$en} & 0.55 & 0.66 & 0.58 & 0.46 & 0.56 \\
     \textbf{WMT'19 fr$\rightarrow$en} & 0.56 & 0.66 & 0.59 & 0.48 & 0.57 \\
     \textbf{WMT'19 en$\rightarrow$zh} & 0.60 & 0.76 & 0.67 & 0.49 & 0.63 \\
     \textbf{HC3-Paraphrase-en} & 0.76 & 0.98 & 0.97 & 0.70 &  0.84 \\
     \textbf{HC3-Paraphrase-zh} & 0.57 & 0.85 & 0.70 & 0.35 &  0.60 \\
     \textbf{HC3-en} & 0.99 & 0.99 & 0.98 & 0.99 &  0.99 \\
     \textbf{HC3-zh} & 0.99 & 0.99 & 0.99 & 0.99 &  0.99 \\
\midrule
\end{tabular}
}
\caption{Experimental results of RoBERTa detectors in Chinese and English respectively. Human and ChatGPT are ground-truth labels for text.}
\label{tabler_one}
\end{table}

We focus on translation, summarization, and paraphrasing datasets to validate the aforementioned observations. We begin by selecting several commonly used human-annotated datasets and inputting their raw sentences into ChatGPT to generate corresponding outputs. To verify whether the detector can effectively distinguish between these types of data, we mix the human-labeled target sentences from the datasets with the sentences generated by ChatGPT. These combined sentences are then provided as input to the detector, which determines whether each text is generated by ChatGPT or written by a human. We use chatgpt-detector-RoBERTa-chinese\footnote{https://huggingface.co/Hello-SimpleAI/chatgpt-detector-roberta-chinese} to detect Chinese and chatgpt-detector-RoBERTa\footnote{https://huggingface.co/Hello-SimpleAI/chatgpt-qa-detector-roberta-chinese} to detect English respectively, as proposed by ~\cite{DBLP:journals/corr/abs-2301-07597}. As shown in Table \ref{tabler_one}, we find that current detectors have difficulty distinguishing the source of sentences. The prediction results are mainly presented in two cases. In the first case, as shown in the translation dataset WMT'19 en$\rightarrow$zh results, the detector performs poorly in terms of precision and recall for both classes. The second case, as shown in the summarization dataset CNN/DailyMail ~\cite{see-etal-2017-get} results, where the detector is always biased towards predicting text as written by a human, exhibits very high recall in the human class but very low recall in the ChatGPT class. For direct comparison with detection results on the QA samples of HC3, we propose HC3-Paraphrase, which contains paraphrases of HC3 questions generated using ChatGPT. Although ChatGPT takes roughly the same input (both question text) for QA and paraphrasing tasks, as shown in Table \ref{tabler_one}, the detection performance on the HC3-Paraphrase is greatly degraded compared to the HC3. The result validates our hypothesis that current detectors struggle to detect semantic-invariant tasks.

In order to fill the gap of HC3 under semantic-invariant tasks, we extend HC3 and propose a larger ChatGPT-generated text dataset covering translation, summarization, and paraphrasing tasks, called HC3 Plus. In addition, language models fine-tuned on a large number of tasks have demonstrated impressive abilities and generalization~\cite{DBLP:conf/emnlp/WangMAKMNADASPK22}. 
% In this paper, we further fine-tune the well-known T\textit{k}-instruct~\cite{DBLP:conf/emnlp/WangMAKMNADASPK22} model to serve as a new baseline for detector models. 
In this paper, to explore the effectiveness of instruction fine-tuning models in detecting machine-generated text, we investigate the use of the T\textit{k}-instruct~\cite{DBLP:conf/emnlp/WangMAKMNADASPK22} model for detecting text generated by ChatGPT.
Our main contribution can be summarized as follows:
\begin{itemize}[leftmargin=*]
    \item We experimentally demonstrate that current detectors cannot detect semantic-invariant samples.
    \item We introduce a more extensive and comprehensive dataset, including semantic-invariant tasks, for ChatGPT-generated text detection.
    \item We use instruction fine-tuning to train a stronger ChatGPT-based text generation detection model.
\end{itemize}

\section{Dataset Construction}
% In this section, we first introduce the datasets involved and then give the data construction process. 

% \subsection{Dataset}
We first select several widely used high-quality corpora that were annotated by humans, encompassing translation, summarization, and paraphrasing tasks.
\begin{itemize}[leftmargin=*]\setlength\itemsep{-0.3em}
    \item The CNN/DailyMail~\cite{see-etal-2017-get} dataset is an English-language dataset containing a large number of unique news articles written by reporters from CNN and the Daily Mail. Each example contains an article and highlights.
    \item The Xsum~\cite{Narayan2018DontGM} is an English-language dataset, covering a wide range of domains, and consists of a BBC article with an accompanying one-sentence summary.
    \item The LCSTS~\cite{hu2015lcsts} is a large-scale Chinese short-text summarization dataset utilizing naturally annotated web resources on Sina Weibo, which is a Chinese social media like Twitter.
    \item  The news2016 corpus, proposed by CLUEbenckmark~\cite{DBLP:conf/coling/XuHZLCLXSYYTDLS20}, is captured from the Chinese We Media (self-media) platform. Each sample contains an article and its corresponding title.
    \item  The WMT translation dataset is a widely used collection of data for machine translation. In this paper, we follow HC3 and mainly focus on the construction of English and Chinese detectors. Therefore, we consider the translation set to include WMT'19 En$\rightarrow$Zh, WMT'19 Zh$\rightarrow$En, WMT'16 Ro$\rightarrow$En, WMT'16 De$\rightarrow$en and WMT'14 Fr$\rightarrow$En.
    \item For the paraphrasing dataset, we choose to use questions proposed by HC3 as the original text of paraphrasing, which can be directly compared with the QA samples proposed by HC3.
\end{itemize}

We use \texttt{GPT-3.5-Turbo-0301} via the OpenAI API to generate target texts. Our proposed dataset, HC3-SI (HC3 Semantic-Invariance), is approximately twice the size of HC3. Details about the composition and size of HC3-SI can be found in Appendix \ref{sec:appendix1}. We then combine HC3-SI with HC3 to create a dataset referred to as HC3 Plus for convenience.
% \begin{figure}[h]
%     \centering
%     \includegraphics[width=0.5\textwidth]{img/fr2en.png}
%     \caption{A gpt-3.5-turbo api call request to translate French to English.}
%     \label{fig:fr2en}
% \end{figure}

% \begin{table}[!ht]
% \scalebox{1}{
% \begin{tabular}{l|c|c|c}
% \multicolumn{4}{l}{\textit{English Dataset}} \\
% \toprule 
% Dataset & train & val  & test \\
% \midrule
%      \textbf{CNN/DailyMail} & 11160 & 1240 & 6000 \\
%      \textbf{Xsum} & 11158 & 1240 & 5994 \\
%      \textbf{WMT'19 de$\rightarrow$en} & 11077 & 1231 & 4264 \\
%      \textbf{WMT'19 fr$\rightarrow$en} & 11043 & 1227 & 5934 \\
%      \textbf{WMT'19 zh$\rightarrow$en} & 11039 & 1227 & 5990 \\
%      \textbf{HC3-Paraphrase-en} & 29233 & 3249 & 6000 \\
% \midrule
%      \textbf{Total in English} & 84710 & 9414 & 34182 \\
% \bottomrule
% \addlinespace
% \multicolumn{4}{l}{\textit{Chinese Dataset}} \\
% \toprule 
% Dataset & train & val  & test \\
% \midrule
%      \textbf{LCSTS} & 11160 & 1240 & 6000 \\
%      \textbf{news2016} & 11023 & 1225 & 5896 \\
%      \textbf{WMT'19 en$\rightarrow$zh} & 10823 & 1203 & 5972 \\
%      \textbf{HC3-Paraphrase-zh} & 9702 & 1078 & 4622 \\
% \midrule
%      \textbf{Total in Chinese} & 42708 & 4746 & 22490 \\
% \bottomrule
% \end{tabular}
% }
% \caption{The size of our proposed dataset.}
% \label{tabler_two}
% \end{table}

\section{Method}
The language model fine-tuned on a large number of instructions shows strong generalization ability on unseen tasks. We propose instructDGGC, to utilize extensive instruction fine-tuning LLM for \textbf{D}etecting Chat\textbf{G}PT-\textbf{G}enerated \textbf{C}ontent. Our method is based on the T\textit{k}-instruct model and further instruction fine-tuning on the HC3 Plus dataset. InstructDGGC is a generative model for discriminating whether an input text is generated by ChatGPT given in-context instructions (task definition and few-shot). 

% First, we follow T\textit{k}-instruct, by instruction-tuning a language model $LM$ on Super-NaturalInstructions ~\cite{DBLP:conf/emnlp/WangMAKMNADASPK22}, which is a multi-task dataset, and get the $LM_{inst}$. 
First, we finetune a Language model on a multi-task dataset, named Super-NaturalInstructions ~\cite{DBLP:conf/emnlp/WangMAKMNADASPK22} to obtain a model $LM_{inst}$.
The obtained $LM_{inst}$ already has a good instruction following ability. Then we further conduct instruction fine-tuning on HC3 Plus. Our task instructions follow the same uniform schema proposed by T\textit{k}-instruct which is composed of the following parts:

\begin{itemize}[leftmargin=*]\setlength\itemsep{-0.3em}
    \item DEFINITION is the description of the task, denoted as $DEF$. In this paper, our current task is to detect whether the text is produced by a machine or not. 
    \item POSITIVE EXAMPLES are input samples and their correct output, denoted as $POS$. In this paper, we give $2$-shot each time, one is a sample generated by ChatGPT, and the other is a sample written by a human.
\end{itemize}
 
For a given instance, we first concatenate it with instructions as input to the model. Then, the model directly outputs the corresponding label (human or model) in an auto-regressive manner. We use the $T$ denoted the input text, the above process can be expressed as:
\begin{equation}
\begin{aligned}
    Out = LM_{Inst}(DEF,POS,T)
\end{aligned}
\end{equation}

\begin{table*}[!ht]
\centering
\scalebox{1}{
\begin{tabular}{l|ccccc}
\multicolumn{6}{l}{\textit{English Dataset}} \\
\toprule
Model & HC3 & Translation & Summarization & Paraphrasing & Overall \\
\midrule
\textbf{RoBERTa-HC3} & 99.40 & 57.79 & 58.49 & 83.03 & 76.81 \\
\textbf{RoBERTa-HC3 Plus} & 98.84 & 71.36 & 99.21 & 96.53 & 89.93 \\
\textbf{InstructDGGC} & \textbf{99.52} & \textbf{75.59} & \textbf{99.49} & \textbf{97.87} & \textbf{91.73} \\
\bottomrule
\addlinespace
\multicolumn{6}{l}{\textit{Chinese Dataset}} \\
\toprule
Model & HC3 & Translation & Summarization & Paraphrasing & Overall \\
\midrule
\textbf{RoBERTa-HC3} & \textbf{98.65} & 62.68 & 53.69 & 59.91 &67.87 \\
\textbf{RoBERTa-HC3 Plus} & 92.64 & 61.49 & 94.25 & \textbf{92.67} & 87.12 \\
\textbf{InstructDGGC} & 92.21 & \textbf{65.19} & \textbf{94.26} & 92.47 & \textbf{87.70}\\
\bottomrule
\end{tabular}
}
\caption{Test set accuracy of the model on the HC3 data and our proposed translation, summarization, paraphrasing data.}
\label{tabler_result}
\end{table*}

\section{Experiment}
First, we follow~\cite{DBLP:journals/corr/abs-2301-07597} to use the \texttt{Roberta-Base}\footnote{https://huggingface.co/FacebookAI/roberta-base} model for training the detector as a baseline. We train two RoBERTa models based on HC3 and HC3 Plus, named RoBERTa-HC3 and RoBERTa-HC3 Plus, respectively. Subsequently, we train the InstructDGGC model, which is divided into two steps. First, we follow the T\textit{k}-instruct to conduct instruction tuning on a large number of tasks, and then further fine-tune the model on HC3 Plus. In this section, we introduce the experimental details. 
% report the corresponding results obtained, and then give the relevant experimental analysis.
\subsection{Experiment Setting}
Before training the detector, we need to perform instruction fine-tuning on a large number of tasks, so that our generative model has good instruction-following ability. For the English model, we use \texttt{T\textit{k}-instruct-base-def-pos}\footnote{https://huggingface.co/allenai/tk-instruct-base-def-pos} released by ~\cite{DBLP:conf/emnlp/WangMAKMNADASPK22}, which has been fine-tuned on 757 English tasks. However, they did not publish a base-size model that can handle Chinese. Therefore, we conduct instruction tuning a language model LM using instruction-equipped data. We used the code released by ~\cite{DBLP:conf/emnlp/WangMAKMNADASPK22}, followed their settings, and trained an instruction fine-tuning model on Super-NaturalInstructions(including 24 Chinese tasks) based on mt5-base ~\cite{DBLP:conf/naacl/XueCRKASBR21}. 

We then further conduct instruction fine-tuning on the HC3 Plus data.
At this stage, for both Chinese and English models, we use the same parameter configuration and we train for $4$ epochs with a learning rate of $1e^{-4}$ and a batch size of $32$. 
% \begin{table}[!ht]
% \centering
% \scalebox{0.83}{
% \begin{tabular}{l|ccc}
% \multicolumn{4}{l}{\textit{English Dataset}} \\
% \toprule
% Model & \multicolumn{3}{c}{HC3-SI Test Scores} \\
% & Precision & Recall & Accuracy \\
% \midrule
% \textbf{RoBERTa-HC3} & 65.48 & 51.55 & 62.19 \\
% \textbf{RoBERTa-HC3 Plus} & 96.60 & 99.90 & 98.84 \\
% \bottomrule
% \addlinespace
% \multicolumn{4}{l}{\textit{Chinese Dataset}} \\
% \toprule
% Model & \multicolumn{3}{c}{HC3-SI Test Scores} \\
% & Precision & Recall & Accuracy \\
% \midrule
% \textbf{RoBERTa-HC3} & 70.75 & 25.05 & 57.35 \\
% \textbf{RoBERTa-HC3 Plus} & 92.64 & 99.84 & 96.62 \\
% \bottomrule
% \end{tabular}
% }
% \caption{Comparison between the model trained on HC3 Plus and the model trained on HC3.}
% \label{tabler_four}
% \end{table}
\subsection{Experiment Result and Analysis}
As shown in Table \ref{tabler_result}, we respectively select the best checkpoint according to the results of the validation sets of HC3 and HC3-SI then give the results of the test set. For HC3-SI, we give the score details of each task, involving translation, summarization, and paraphrasing tasks. 

First, we focus on the RoBERTa-based model. RoBERTa-HC3 performed poorly on our proposed dataset, indicating that the model struggles to handle semantic-invariant data generated by ChatGPT. We then used HC3 Plus to train RoBERTa. For semantic-invariant data, we observed that, aside from a decline in performance on the Chinese translation data, the model's performance on other datasets improved significantly.
We noted that for the translation task, ChatGPT generates sentences that are much more similar to the target sentences in the dataset compared to tasks like summarization and paraphrasing. As a result, detecting AI-generated content in translation data is more challenging for the model, even with a substantial amount of training data. Appendix \ref{sec:appendix3} provides examples and a more detailed analysis.
Additionally, we observed that RoBERTa-HC3, when trained only on HC3 data, demonstrates stronger detection capabilities on HC3 compared to RoBERTa-HC3 Plus. This suggests that training a detector for a specific task is easier than training one that needs to accommodate various types of task data.

Next, we focus on InstructDGGC. As shown in Table \ref{tabler_result}, we observe a significant improvement in performance on English data compared to RoBERTa-HC3 Plus, demonstrating the strong generalization capabilities of language models fine-tuned on a large number of instructions. However, for Chinese data, aside from a noticeable improvement in translation tasks, the results for other tasks remain largely unchanged. We believe the limited improvement is due to the model's insufficient instruction-following capability, likely stemming from a lack of fine-tuning data for Chinese instructions. Moreover, compared to HC3 Plus, InstructDGGC performs better on translation data, further confirming that it is more robust.

Finally, we focus on the overall score. Compared to RoBERTa-HC3 Plus, InstructDGGC achieved better overall performance, with an improvement of $1.8$\% for English data and $0.58$\% for Chinese data. This indicates that fine-tuning instruction models for detecting generated text is a promising method.

\section{Conclusion}
In this work, we experimentally demonstrate that detecting ChatGPT-generated text becomes more difficult on semantic-invariant tasks, for which we propose a more extensive and comprehensive dataset HC3 Plus. And we use the instruction fine-tuning model to achieve a detector and experimental results show that our detector outperforms the RoBERTa-based method proposed by previous work. We hope that our analysis of semantic-invariant text detection can provide some insights for future work.

\section{Limitations}
While our experiments show that detecting semantic-invariant text is more challenging for detectors, our dataset only considers the current version of ChatGPT, i.e., \texttt{GPT-3.5-Turbo-0301}. As ChatGPT undergoes continuous iterative improvement, our dataset may become outdated compared to the evolving versions of ChatGPT. In future work, we plan to further investigate the impact of these iterations on detection performance.

% \newpage
\bibliography{anthology,custom}

\begin{thebibliography}{13}
\expandafter\ifx\csname natexlab\endcsname\relax\def\natexlab#1{#1}\fi

\bibitem[{Gehrmann et~al.(2019)Gehrmann, Strobelt, and Rush}]{DBLP:conf/acl/GehrmannSR19}
Sebastian Gehrmann, Hendrik Strobelt, and Alexander~M. Rush. 2019.
\newblock \href {https://doi.org/10.18653/v1/p19-3019} {{GLTR:} statistical detection and visualization of generated text}.
\newblock In \emph{Proceedings of the 57th Conference of the Association for Computational Linguistics, {ACL} 2019, Florence, Italy, July 28 - August 2, 2019, Volume 3: System Demonstrations}, pages 111--116. Association for Computational Linguistics.

\bibitem[{Guo et~al.(2023)Guo, Zhang, Wang, Jiang, Nie, Ding, Yue, and Wu}]{DBLP:journals/corr/abs-2301-07597}
Biyang Guo, Xin Zhang, Ziyuan Wang, Minqi Jiang, Jinran Nie, Yuxuan Ding, Jianwei Yue, and Yupeng Wu. 2023.
\newblock \href {https://doi.org/10.48550/arXiv.2301.07597} {How close is chatgpt to human experts? comparison corpus, evaluation, and detection}.
\newblock \emph{CoRR}, abs/2301.07597.

\bibitem[{Hu et~al.(2015)Hu, Chen, and Zhu}]{hu2015lcsts}
Baotian Hu, Qingcai Chen, and Fangze Zhu. 2015.
\newblock Lcsts: A large scale chinese short text summarization dataset.
\newblock \emph{arXiv preprint arXiv:1506.05865}.

\bibitem[{Ippolito et~al.(2020)Ippolito, Duckworth, Callison{-}Burch, and Eck}]{DBLP:conf/acl/IppolitoDCE20}
Daphne Ippolito, Daniel Duckworth, Chris Callison{-}Burch, and Douglas Eck. 2020.
\newblock \href {https://doi.org/10.18653/v1/2020.acl-main.164} {Automatic detection of generated text is easiest when humans are fooled}.
\newblock In \emph{Proceedings of the 58th Annual Meeting of the Association for Computational Linguistics, {ACL} 2020, Online, July 5-10, 2020}, pages 1808--1822. Association for Computational Linguistics.

\bibitem[{Liu et~al.(2019)Liu, Ott, Goyal, Du, Joshi, Chen, Levy, Lewis, Zettlemoyer, and Stoyanov}]{DBLP:journals/corr/abs-1907-11692}
Yinhan Liu, Myle Ott, Naman Goyal, Jingfei Du, Mandar Joshi, Danqi Chen, Omer Levy, Mike Lewis, Luke Zettlemoyer, and Veselin Stoyanov. 2019.
\newblock \href {http://arxiv.org/abs/1907.11692} {Roberta: {A} robustly optimized {BERT} pretraining approach}.
\newblock \emph{CoRR}, abs/1907.11692.

\bibitem[{Mitchell et~al.(2023)Mitchell, Lee, Khazatsky, Manning, and Finn}]{DBLP:journals/corr/abs-2301-11305}
Eric Mitchell, Yoonho Lee, Alexander Khazatsky, Christopher~D. Manning, and Chelsea Finn. 2023.
\newblock \href {https://doi.org/10.48550/arXiv.2301.11305} {Detectgpt: Zero-shot machine-generated text detection using probability curvature}.
\newblock \emph{CoRR}, abs/2301.11305.

\bibitem[{Narayan et~al.(2018)Narayan, Cohen, and Lapata}]{Narayan2018DontGM}
Shashi Narayan, Shay~B. Cohen, and Mirella Lapata. 2018.
\newblock Don't give me the details, just the summary! topic-aware convolutional neural networks for extreme summarization.
\newblock \emph{ArXiv}, abs/1808.08745.

\bibitem[{Radford et~al.(2019)Radford, Wu, Child, Luan, Amodei, Sutskever et~al.}]{radford2019language}
Alec Radford, Jeffrey Wu, Rewon Child, David Luan, Dario Amodei, Ilya Sutskever, et~al. 2019.
\newblock Language models are unsupervised multitask learners.
\newblock \emph{OpenAI blog}, 1(8):9.

\bibitem[{Sadasivan et~al.(2023)Sadasivan, Kumar, Balasubramanian, Wang, and Feizi}]{DBLP:journals/corr/abs-2303-11156}
Vinu~Sankar Sadasivan, Aounon Kumar, Sriram Balasubramanian, Wenxiao Wang, and Soheil Feizi. 2023.
\newblock \href {https://doi.org/10.48550/arXiv.2303.11156} {Can ai-generated text be reliably detected?}
\newblock \emph{CoRR}, abs/2303.11156.

\bibitem[{See et~al.(2017)See, Liu, and Manning}]{see-etal-2017-get}
Abigail See, Peter~J. Liu, and Christopher~D. Manning. 2017.
\newblock \href {https://doi.org/10.18653/v1/P17-1099} {Get to the point: Summarization with pointer-generator networks}.
\newblock In \emph{Proceedings of the 55th Annual Meeting of the Association for Computational Linguistics (Volume 1: Long Papers)}, pages 1073--1083, Vancouver, Canada. Association for Computational Linguistics.

\bibitem[{Wang et~al.(2022)Wang, Mishra, Alipoormolabashi, Kordi, Mirzaei, Naik, Ashok, Dhanasekaran, Arunkumar, Stap, Pathak, Karamanolakis, Lai, Purohit, Mondal, Anderson, Kuznia, Doshi, Pal, Patel, Moradshahi, Parmar, Purohit, Varshney, Kaza, Verma, Puri, Karia, Doshi, Sampat, Mishra, A, Patro, Dixit, and Shen}]{DBLP:conf/emnlp/WangMAKMNADASPK22}
Yizhong Wang, Swaroop Mishra, Pegah Alipoormolabashi, Yeganeh Kordi, Amirreza Mirzaei, Atharva Naik, Arjun Ashok, Arut~Selvan Dhanasekaran, Anjana Arunkumar, David Stap, Eshaan Pathak, Giannis Karamanolakis, Haizhi~Gary Lai, Ishan Purohit, Ishani Mondal, Jacob Anderson, Kirby Kuznia, Krima Doshi, Kuntal~Kumar Pal, Maitreya Patel, Mehrad Moradshahi, Mihir Parmar, Mirali Purohit, Neeraj Varshney, Phani~Rohitha Kaza, Pulkit Verma, Ravsehaj~Singh Puri, Rushang Karia, Savan Doshi, Shailaja~Keyur Sampat, Siddhartha Mishra, Sujan~Reddy A, Sumanta Patro, Tanay Dixit, and Xudong Shen. 2022.
\newblock \href {https://aclanthology.org/2022.emnlp-main.340} {Super-naturalinstructions: Generalization via declarative instructions on 1600+ {NLP} tasks}.
\newblock In \emph{Proceedings of the 2022 Conference on Empirical Methods in Natural Language Processing, {EMNLP} 2022, Abu Dhabi, United Arab Emirates, December 7-11, 2022}, pages 5085--5109. Association for Computational Linguistics.

\bibitem[{Xu et~al.(2020)Xu, Hu, Zhang, Li, Cao, Li, Xu, Sun, Yu, Yu, Tian, Dong, Liu, Shi, Cui, Li, Zeng, Wang, Xie, Li, Patterson, Tian, Zhang, Zhou, Liu, Zhao, Zhao, Yue, Zhang, Yang, Richardson, and Lan}]{DBLP:conf/coling/XuHZLCLXSYYTDLS20}
Liang Xu, Hai Hu, Xuanwei Zhang, Lu~Li, Chenjie Cao, Yudong Li, Yechen Xu, Kai Sun, Dian Yu, Cong Yu, Yin Tian, Qianqian Dong, Weitang Liu, Bo~Shi, Yiming Cui, Junyi Li, Jun Zeng, Rongzhao Wang, Weijian Xie, Yanting Li, Yina Patterson, Zuoyu Tian, Yiwen Zhang, He~Zhou, Shaoweihua Liu, Zhe Zhao, Qipeng Zhao, Cong Yue, Xinrui Zhang, Zhengliang Yang, Kyle Richardson, and Zhenzhong Lan. 2020.
\newblock \href {https://doi.org/10.18653/v1/2020.coling-main.419} {{CLUE:} {A} chinese language understanding evaluation benchmark}.
\newblock In \emph{Proceedings of the 28th International Conference on Computational Linguistics, {COLING} 2020, Barcelona, Spain (Online), December 8-13, 2020}, pages 4762--4772. International Committee on Computational Linguistics.

\bibitem[{Xue et~al.(2021)Xue, Constant, Roberts, Kale, Al{-}Rfou, Siddhant, Barua, and Raffel}]{DBLP:conf/naacl/XueCRKASBR21}
Linting Xue, Noah Constant, Adam Roberts, Mihir Kale, Rami Al{-}Rfou, Aditya Siddhant, Aditya Barua, and Colin Raffel. 2021.
\newblock \href {https://doi.org/10.18653/v1/2021.naacl-main.41} {mt5: {A} massively multilingual pre-trained text-to-text transformer}.
\newblock In \emph{Proceedings of the 2021 Conference of the North American Chapter of the Association for Computational Linguistics: Human Language Technologies, {NAACL-HLT} 2021, Online, June 6-11, 2021}, pages 483--498. Association for Computational Linguistics.

\end{thebibliography}
\bibliographystyle{acl_natbib}
% \newpage
\appendix
\section{Related Work}
\label{sec:appendix2}
OpenAI released a neural network-based detector to facilitate the detection of outputs of GPT-2 ~\cite{radford2019language} models. Another stream of work focuses on zero-shot AI text detection without any additional training overhead ~\cite{DBLP:journals/corr/abs-2301-11305, DBLP:conf/acl/IppolitoDCE20, DBLP:conf/acl/GehrmannSR19}. After the release of ChatGPT, ~\cite{DBLP:journals/corr/abs-2301-07597} collected a human ChatGPT comparison corpus HC3. Then based on this dataset, they trained detection systems to distinguish whether the text was generated by ChatGPT or humans. However, HC3 mainly focuses on the question-answering task, which neglects the semantic-invariant tasks popular in real-world scenarios. ~\cite{DBLP:journals/corr/abs-2303-11156} found that using a lightweight paraphraser can cause most detectors to fail. Furthermore, ~\cite{DBLP:conf/emnlp/WangMAKMNADASPK22} proposed T\textit{k}-instruct, which performed fine-tuning on a large number of instructions and showed strong generalization ability. To counter the above issues, we propose a new dataset that considers semantic-invariant scenarios. We also train a novel detector based on T\textit{k}-instruct which achieves better detection performance.

\section{HC3-SI Dataset Description}
As shown in Table \ref{tabler_four}, we give the number of samples in the data set HC3-SI. For Chinese data, the number of samples in the train/val/test set are $42708/4746/22516$. For English data, the number of samples in the train/val/test set The sample sizes are $95745/10641/38142$ respectively. The total sample size of our proposed HC3-SI is $210,000$, nearly twice that of HC3. In addition, our data does not overlap with HC3, and they can be used together.
\label{sec:appendix1}
\begin{table}[!ht]
\scalebox{0.8}{
\begin{tabular}{l|c|c|c}
\multicolumn{4}{l}{\textit{English Dataset}} \\
\toprule 
Dataset & train & val  & test \\
\midrule
     \textbf{CNN/DailyMail} & 11160 & 1240 & 6000 \\
     \textbf{Xsum} & 11158 & 1240 & 5994 \\
     \textbf{WMT'19 de$\rightarrow$en} & 11077 & 1231 & 4264 \\
     \textbf{WMT'19 fr$\rightarrow$en} & 11043 & 1227 & 5934 \\
     \textbf{WMT'19 zh$\rightarrow$en} & 11039 & 1227 & 5990 \\
     \textbf{WMT'19 ro$\rightarrow$en} & 11035 & 1227 & 3960 \\
     \textbf{HC3-Paraphrase-en} & 29233 & 3249 & 6000 \\
\midrule
     \textbf{Total in English} & 95745 & 10641 & 38142 \\
\bottomrule
\addlinespace
\multicolumn{4}{l}{\textit{Chinese Dataset}} \\
\toprule 
Dataset & train & val  & test \\
\midrule
     \textbf{LCSTS} & 11160 & 1240 & 6000 \\
     \textbf{news2016} & 11023 & 1225 & 5918 \\
     \textbf{WMT'19 en$\rightarrow$zh} & 10823 & 1203 & 5976 \\
     \textbf{HC3-Paraphrase-zh} & 9702 & 1078 & 4622 \\
\midrule
     \textbf{Total in Chinese} & 42708 & 4746 & 22516 \\
\bottomrule
\end{tabular}
}
\caption{The size of our proposed dataset HC3-SI.}
\label{tabler_four}
\end{table}

\section{Sample Description}
\label{sec:appendix3}
\begin{figure*}[h]
    \centering
    \includegraphics[width=1\textwidth]{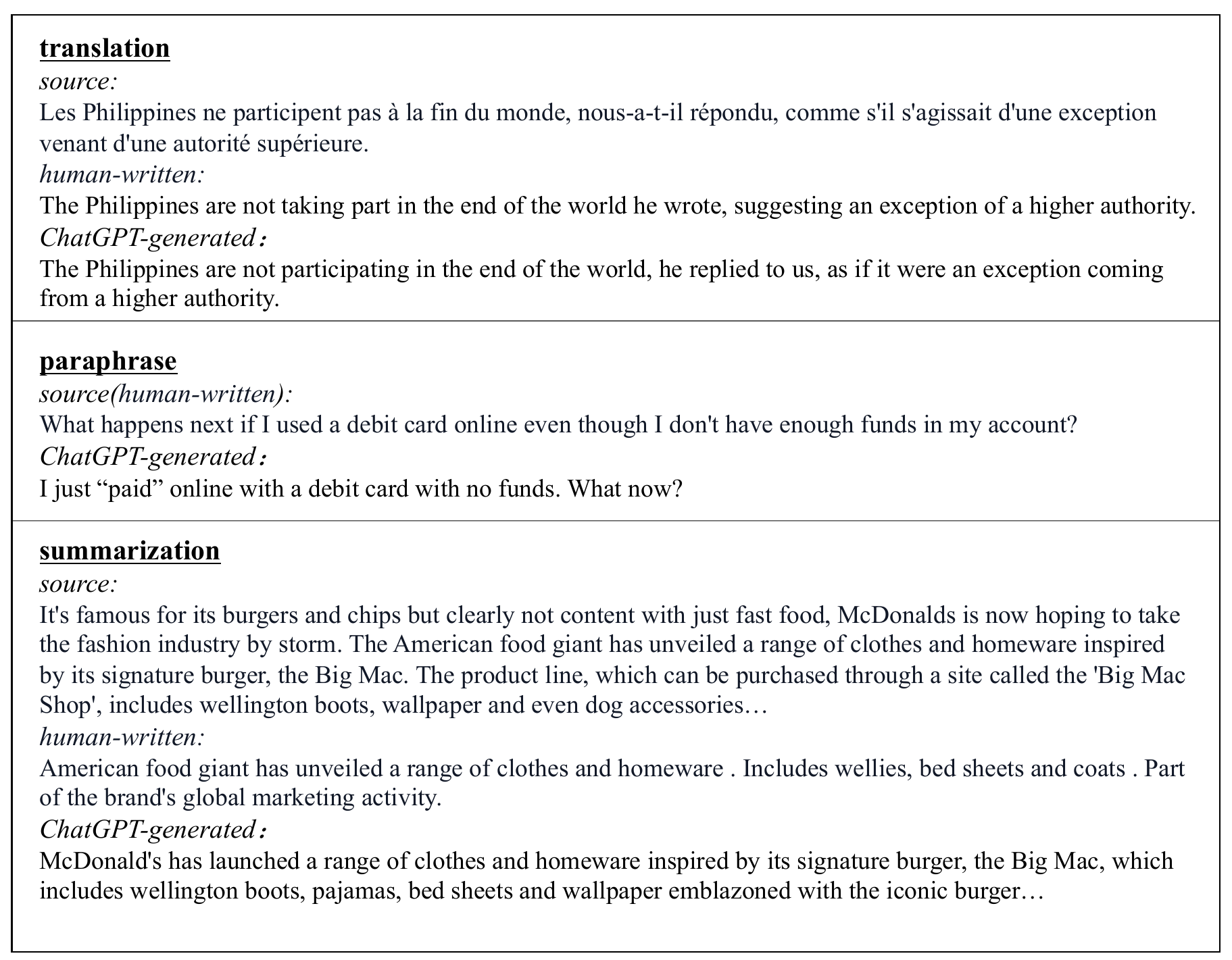}
    \caption{Examples of translation, summary generation, and paraphrasing in HC3-SI}
    \label{fig:samples}
\end{figure*}

Figure \ref{fig:samples} gives examples of translation, summary generation, and paraphrasing respectively in HC3-SI. Compared with summarization and paraphrasing, the translation task has stronger semantic constraints from the source sentence, which leads to a higher similarity between the text generated by ChatGPT and the text written by humans. Taking the French-to-English sample in Figure \ref{fig:samples} as an example, the ChatGPT-generated text has a high overlap of word fragments and many synonymous replacement words compared with the human-written text. Such samples are indiscriminate sources for detectors, which is also verified by our experimental results. For summarization and paraphrasing, although the text generated by ChatGPT and the text written by humans have roughly the same semantics, there are obvious differences in style. Therefore, a well-trained detector will recognize them more easily.

\end{document}